\documentclass[conference]{IEEEtran}
\IEEEoverridecommandlockouts
\usepackage{cite}
\usepackage{soul}
\usepackage{amsmath,amssymb,amsfonts}
\usepackage{algorithm}
\usepackage{algorithmic}
\usepackage[utf8]{inputenc}
\usepackage{multirow}
\usepackage{array}
\usepackage{lipsum}
\usepackage{subcaption}
\usepackage{textcomp}
\usepackage{xcolor}
\usepackage{graphicx}
\usepackage{textcomp}
\usepackage{amsmath}
\usepackage[left=0.625in, right=0.625in, top=0.75in, bottom=1in, footskip=0.25in]{geometry}
\usepackage{tikz}
\usepackage{amssymb}
\usepackage{graphicx}
\usepackage{cite}
\usepackage{bm}
\usepackage{dsfont}
\usepackage{graphics} 
\usepackage{epsfig} 
\usepackage{stackrel}
\usepackage{geometry}{}
\usepackage{float}
\usepackage{url}
\def\BibTeX{{\rm B\kern-.05em{\sc i\kern-.025em b}\kern-.08em
    T\kern-.1667em\lower.7ex\hbox{E}\kern-.125emX}}

\begin{document}

\title{\textbf{PILOT}: High-\textbf{P}recision \textbf{I}ndoor \textbf{Lo}calization for Au\textbf{t}onomous Drones}

\author{\IEEEauthorblockN{Alireza Famili, Angelos Stavrou, Haining Wang, Jung-Min (Jerry) Park}
	\IEEEauthorblockA{\textit{Department of Electrical and Computer Engineering, Virginia Tech} \\
		\{afamili, angelos, hnw, jungmin\}@vt.edu}
}

\maketitle

\begin{abstract}
In many scenarios, unmanned aerial vehicles (UAVs), aka drones, need to have the capability of autonomous flying to carry out their mission successfully. In order to allow these autonomous flights, drones need to know their location constantly. Then, based on the current position and the final destination, navigation commands will be generated and drones will be guided to their destination. Localization can be easily carried out in outdoor environments using GPS signals and drone inertial measurement units (IMUs). However, such an approach is not feasible in indoor environments or GPS-denied areas. In this paper, we propose a localization scheme for drones called PILOT (High-Precision Indoor Localization for Autonomous Drones) that is specifically designed for indoor environments. PILOT relies on ultrasonic acoustic signals to estimate the target drone’s location. In order to have a precise final estimation of the drone’s location, PILOT deploys a three-stage localization scheme. The first two stages provide robustness against the multi-path fading effect of indoor environments and mitigate the ranging error. Then, in the third stage, PILOT deploys a simple yet effective technique to reduce the localization error induced by the relative geometry between transmitters and receivers and significantly reduces the height estimation error. The performance of PILOT was assessed under different scenarios and the results indicate that PILOT achieves centimeter-level accuracy for three-dimensional localization of drones.
\end{abstract}

\begin{IEEEkeywords}
indoor localization, drones, ultrasound transceiver, autonomous flying 
\end{IEEEkeywords}

\section{Introduction}
\noindent Over the past few years, the global drone industry has grown progressively and the number of applications in which civilian drones play an essential role in both indoor and outdoor environments has therefore increased. There are a wide range of indoor drone applications today, ranging from recreational use cases to essential safety-of-life use cases. Examples include shipping and delivery of packages, aerial photography for capturing footage and geographical mapping, providing temporary cellular coverage in case of disasters, reconnaissance inside nuclear power plants, helping firefighters locate individuals inside burning buildings, and security surveillance inside large building complexes among others.    

Drones need to fly fully or partially autonomously in all of the above applications to carry out their mission successfully. To allow such fully or partially autonomous flights, the ground control station or any other infrastructure that supports the operation of the drone needs to continuously localize and monitor the drone's position and send this information to the navigation controller of the drone to provide the capability of autonomous navigation for the drone. Localization and monitoring the drone's position can be easily carried out in outdoor environments using GPS and the Inertial Measurement Units (IMUs) of the drone without seeking aid from the ground controller or any other extra infrastructure to perform this task. Such an approach is, however, not feasible in indoor environments or GPS-denied areas. 

Vision-based approaches are widely used for drone localization when GPS is not accessible (e.g.,~\cite{Novel_Visual_Odometry}). However, because of the vibration of the drone during flight, the accuracy of current vision-based methods is generally limited. In addition, in vision-impaired conditions (e.g., low light conditions or blockage of the line of sight), the accuracy will further deteriorate. In addition to all the mentioned issues, vision-based approaches are expensive both in terms of the hardware cost and the computational complexity. 

In addition to vision-based methods, there are other techniques, including, Doppler-shift-based tracking (e.g.,~\cite{Potential_Sound_Based}), fingerprinting-based localization using received signal strength (RSS) or channel state information (CSI) (e.g.,~\cite{Automating_CSI_Measurement,DL_RSS_CSI,Fusion_CSI_RSS}), and localization using cellular networks (e.g.,~\cite{Signal_of_Opportunity}). Doppler-shift-based tracking has limitations; by itself, it does not provide sufficient tracking accuracy. Fingerprinting-based techniques are too vulnerable to any changes that happen in the environment after the off-line phase of collecting the RSS or CSI information; hence, they are not capable of providing high accuracy localization. With the current technology, localization using cellular networks has accuracy in the order of tens of meters and cannot provide centimeter-level high accuracy localization. In addition, there is a good chance of having weak cellular coverage in indoor environments, which itself degrades the accuracy drastically.

Another well-known class for localization in the absence of a GPS signal is the ranging-based method. This category uses the concept behind GPS and makes an indoor positioning system that resembles how localization using GPS works, i.e., there are some fixed beacons in the room (similar to the satellites) and the localization task gets done by performing some measurement techniques on the communicated signals between these beacons and the target object. Ranging-based methods either use Radio Frequency (RF) signal (RF-based localization, e.g.,~\cite{Freq_Hopping_WiFi}) or they are working based on acoustic signals (acoustic-based localization, e.g.,~\cite{OFDM_Acoustic_ToF}). The major problem with ranging-based methods is the degradation of localization accuracy due to (i) multi-path fading effect and (ii) the relative geometry between the transmitter(s) and receiver(s).

In this paper, we propose a three-dimensional localization scheme for drones in GPS-denied indoor environments that is referred to as \emph{High-\textbf{P}recision \textbf{I}ndoor \textbf{Lo}calization for Au\textbf{t}onomous Drones (PILOT)}. PILOT is based on the ranging-based localization approach with significant enhancement to provide centimeter-level localization accuracy for drones in indoor environments. PILOT uses ultrasound acoustic-based signals for localization. We claim that localization based on acoustic signals provides a variety of advantages over approaches based on RF. Compared to RF signals, the significantly slower propagation speed of acoustic signals allows for higher localization accuracy without the need for having expensive equipment with a high sampling rate. PILOT uses high-frequency acoustic signals, referred to as \emph{ultrasound}, to prevent any interference with the propeller-generated noise of the drone or human-generated noise in the environment. 

Despite most of the available indoor positioning schemes that focus merely on two-dimensional localization, PILOT not only offers localization in three dimensions, but it also proposes additional solution to alleviate the estimation error in Z-axis. Moreover, to provide an accurate location estimation for drones, PILOT leverages some techniques to counter the multi-path fading effect of indoor environments. To the best of our knowledge, PILOT is the first to make completely different and innovative use of these techniques and proposes novel improvements on them to provide a three-dimensional positioning scheme that is robust against the multi-path fading effect and offers centimeter-level accuracy for location estimation of drones in indoor environments without any dependency on GPS signals. The summary of our contributions is as follows. 

$\bullet$ We propose PILOT, a high-accuracy three-dimensional localization scheme for drones in indoor environments that is highly resilient to noise and multi-path fading and achieves high accuracy by employing three stages.

$\bullet$ PILOT overcomes the multi-path fading effect of the indoor environments by employing the Frequency Hopping Spread Spectrum (FHSS) technique for signal communication, which to the best of our knowledge, is a novel approach for localizing a drone in three-dimensional space. 

$\bullet$ PILOT mitigates the localization error and provides further accuracy compared to the merely ranging-based schemes by leveraging the Doppler-shift effect to measure the velocity of the drone and designs a Kalman filter for velocity and distance incorporation.  

$\bullet$ PILOT leverages an additional ultrasound transceiver and uses the reflected ultrasonic signal bounced from the ceiling to measure the drone's height separately and combine it with the available height measurements from the first two stages in a filter to improve the Z-axis estimation accuracy.

$\bullet$ Our comprehensive simulation and experimental results indicate that PILOT achieves a three-dimensional localization error of less than $1.2$~cm. Also, it significantly reduces the Z-axis estimation error compared to the prior state-of-the-art.

The remainder of this paper is structured as follows. In the next section, we will review some of the relevant works in this area. Then, in section~\ref{sec:Robust FHSS localization}, \ref{sec:Enhancing the Robustness of FHSS localization and Tracking}, and~\ref{sec:Trilateration}, which are the core sections of our paper, we will thoroughly explain our scheme for localizing a drone in indoor environments. Followed by these sections, in section~\ref{sec:Preliminary Simulations}, we will describe our simulation test setup and showcase the preliminary performance assessment results on PILOT. Next, in section~\ref{sec:Height Enhancement}, we will be investigating the reason behind having a bad $Z$-axis estimation compared to the $X-Y$ plane and we will be providing our solution to rectify this issue. Next, in section~\ref{sec:Experiments}, first, we describe our experimental test setup and then provide the final evaluation results on PILOT. Finally, in section~\ref{sec:Conclusion}, we conclude our work.

\section{Related Work} \label{sec:Related Work}

Our work is related to the following research areas: (i) indoor localization, (ii) tracking the trajectory of a moving object, and (iii) autonomous navigation of UAVs in the absence (or lack) of GPS signals.

Indoor localization has been a topic of interest for decades~\cite{Indoor_Localization_New_1,Wireless_Acoustic_Sensor_Network,Low_Cost_Ultrasonic,Ultrasound_Wireless_Sensor_Network,OFDM_Acoustic_ToF,Acoustic_Image_Indoor_Localization,Ultrasonic_Chirps,FSK}. The two most popular approaches for indoor localization are ranging-based methods and fingerprinting. In the latter one, localization is based on the comparison of the received signal with an already available offline map of the indoor environment and it can be the map of the CSI or the RSS for the indoor environment~\cite{Automating_CSI_Measurement,DL_RSS_CSI,Fusion_CSI_RSS,RSSI_New_1}. The main drawback of the fingerprinting method is the vulnerability to any changes in the online phase which has not been seen in the offline map and which degrades accuracy.

In the ranging-based localization methods, there is a signal transmission between the target object and multiple receivers at known locations~\cite{Freq_Hopping_WiFi,Spread_Spectrum_and_MEMS,Robust_Broadband,Indoor_Map_Acoustic}. In this category, localization occurs by applying techniques such as trilateration or angulation on the measured distances between the transmitter and all the receivers. The distance calculation is done by measuring some of the features of the received signal, such as time of arrival (TOA) or angle of arrival (AOA). The signal to be communicated in this method can be RF, acoustic, or ultrasound. In~\cite{Freq_Hopping_WiFi}, Chen et al. used a WiFi platform to achieve centimeter-level accuracy for indoor localization. However, RF signals propagate at the speed of light, which makes the accurate receiving process hard and expensive both in terms of computation and hardware cost.

In terms of tracking the trajectory of a moving object, it is essential that the system has the capability to localize the object continuously while it is in motion. There are research works on this topic~\cite{MobiSys_Follow_Me_Drone,Acoustic_Gesture_Tracking,Array_Track,mTrack,SwordFight,FingerIO,RF_IDraw,RFID_Tags,WiDraw,Mouse_in_Air,ToneTrack,CAT}. As an example, in~\cite{MobiSys_Follow_Me_Drone}, Mao et al. used acoustic signals for tracking drones. To make the system robust against the multi-path fading effect of the indoor environment, they were leveraging frequency modulated continuous wave (FMCW) signals and to further enhance the accuracy of the tracking process, they were using a Kalman filter to incorporate both the velocity and distance estimation. The problem with the acoustic signal is that human-generated noise or the drone's propeller noise may interfere with the signal, hence degrading localization accuracy. The other drawback of their work is that their system is just for tracking the drone on a line (one dimension); however, in reality, we need to handle the three-dimensional localization and tracking. Another research work presented sound-based direction sensing where simply by using the shift in the received signal's frequency (Doppler-shift effect), they found the direction of the drone, but again there is no exact localization here~\cite{Potential_Sound_Based}.

Finally, for autonomous navigation of UAVs in indoor environments, there are several methods that research groups have been working on, including: vision-based models using different visual techniques such as visual odometry (VO), simultaneous localization and mapping (SLAM), and optical flow technique~\cite{Novel_Visual_Odometry,low_cost_solution,6_Dimensional,Survey_UAV_navigation_GPS_denied,Vision_Based_New_1,Vision_Based_New_2,Vision_Based_New_3,Vision_Based_New_4,Vision_Based_New_5}. There are also a few research papers that showcase the use of deep neural networks in combination by visual techniques (e.g.,~\cite{Neural_Network}) or use of LiDAR for autonomous flying (e.g.,~\cite{LiDAR}). In another research study~\cite{ROLATIN}, Famili et al. proposed a revised ranging technique for localization of drones in indoor environments. Even though they claimed that their proposed scheme offers high accuracy localization, their work still has room for improvement. Firstly, they did not provide any real-life experimental tests using actual drones to support their localization scheme. Moreover, in their scheme, the $Z$-axis localization error is drastically more than the localization error for the $X-Y$ plane, which degrades the overall three-dimensional accuracy. They did not propose a solution to resolve this issue. Lastly, they assumed that ultrasound sensors have omni-directional transmission capability, thinking that, just one ultrasound transmitter aboard the drone can send signals to all the receivers in the room. However, this is not a fair assumption in real-world scenarios because ultrasound transmitters are not omni-directional. Ultrasound transmitters have a narrow beam propagation pattern; hence, the transmitting signals of the ultrasound transmitter on the drone can only be received by an ultrasound receiver located in that propagation range, not by all the receivers in the room. To address these issues, our scheme proposes a solution to fix the $Z$-axis localization error. Moreover, we take the restrictions of actual ultrasound sensors into account and propose a multi-transmitter system that can compensate for the narrow beam propagation pattern of ultrasound sensors. Finally, we conduct simulations coupled with real-life experiments using an actual drone to evaluate our proposed scheme.

\section{Robust Ranging using FHSS Ultrasound Signals} \label{sec:Robust FHSS localization}
To accomplish fully or partially autonomous drone navigation, the most critical task is constant localization of the drone in three dimensions with high accuracy. We explain how PILOT performs a distance estimation that is highly robust against the noise and the multi-path fading effect of an indoor environment in this section, and in the next two sections, we will discuss how it further improves the accuracy of distance estimation and performs the three-dimensional localization.

As discussed earlier, the ranging-based method is an excellent alternative for localization in the absence of GPS signals. To briefly explain this class of localization, it is required to know the three major concepts: (i) signals to be measured for ranging, (ii) the measurement method, and (iii) the localization technique. Examples of signals deployed for the measurements are RF, acoustics, ultrasound, visible light, and others. Well-known measurement methods for ranging-based localization include angle of arrival (AOA), time of arrival (TOA), time difference of arrival (TDOA), received signal strength (RSS), etc. Finally, techniques for location estimation are angulation, lateration, and fingerprinting. 

RF signals propagate in the speed of light (approximately $3\times10^8~(m/s)$), which is orders of magnitude faster than the acoustics or ultrasound which they travel in the speed of sound (approximately $340~(m/s)$); therefore, for having a high accuracy localization, RF-based systems require more expensive equipment with high sampling rate. In addition, RF signals travel through walls and ceilings, causing interference with the localization system. Finally, there are places (e.g., hospitals, military bases, etc.) where RF signals are prohibited. All being said, to provide high accuracy localization with the minimum hardware and computational cost, PILOT leverages the acoustic waveform as the signal to measure. In addition, to avoid any interference with human or propeller-generated noise in the audible ranges of acoustic signals, PILOT uses the higher frequencies known as \emph{ultrasound}.  

AOA and angulation require special antenna arrays and applying some computationally complicated techniques such as the multiple signal classification (MUSIC)~\cite{MobiSys_Follow_Me_Drone} which incurs high complexity calculations and makes the approach expensive both in terms of hardware cost and processing power. Merely RSS, in which the distances between the target object and the beacons are calculated based on the received power and the path loss formulas, suffer from poor accuracy. To solve this, RSS-fingerprinting methods in which the localization takes place in two steps are offered. In the first step, called the offline phase, an RSS map of the entire venue is collected. Then, in the online phase, localization takes place by comparing the measured received power with the data from the offline phase. Although this improves localization accuracy compared to RSS itself, it is too sensitive to real-time changes; therefore, it is not reliable for high accuracy localization.

PILOT uses TOA of received ultrasound signals for ranging. Compared to other measurement methods, TOA seems to be the best choice in terms of accuracy, computation's simplicity, and implementation cost. The main challenge of relying on the TOA of the received signal is the precise time of arrival detection. In indoor environments, multi-path fading is an inevitable effect due to the reflection of the original signal from walls, ceiling, floor, and other artifacts within the room. Multi-path fading may be a significant problem for measuring the precise TOA, as several copies of the original signal with varying arrival times make it difficult to detect the exact TOA of the original signal. To minimize the multi-path effect and be able to detect the exact TOA of the original signal, PILOT leverages the FHSS technique for signal transmission.

FHSS is a well-known technique where uses different frequency hops as carrier frequencies and spreads the signal. It has applications both in military transmission mainly to avoid jamming and also civilian use, such as the well-known Bluetooth, mainly to provide multi-user capability. We are applying this spread spectrum technique to work in our favor, i.e., we are using the hopping over different frequencies as a method to overcome multi-path fading. To the best of our knowledge, this is the first time using the FHSS technique to counter the multi-path fading effect and offer more accurate location estimation for drones.

There are two kinds of frequency hopping techniques, slow frequency hopping (SFH) and fast frequency hopping (FFH). In SFH, one or more data bits are transmitted within one hop. An advantage is that coherent data detection is possible. In the latter kind, one data bit is divided over multiple hops and coherent signal detection is complex and seldom used. PILOT uses the SFH.

Similar to the approach used in~\cite{Robust_Broadband}, in our scheme, the transmitting signal of the $k$-th drone is modulated using Binary Phase Shift Keying (BPSK) modulation and then it is spread using a sinusoidal signal with variable frequencies depending on the pseudo-random code:
\begin{eqnarray}\label{Tx_FHSSS}
s^{(k)}(t) = d^{(k)}\cdot pT_B(t)\cdot \sin(2\pi f_mt+\phi),
\end{eqnarray}
where $d^{(k)}$ is the transmitted data symbol of the ultrasonic transmitter on the $k$-th drone, the rectangular pulse $pT_B$ is equal to $1$ for $0 \leq t < T_B$ and zero otherwise, $T_B$ is the data symbol duration, and $f_m$ is the set of frequencies over which the signal hops. Then the received signal is in the form of:
\begin{equation}
\begin{split}
r^{(k)} = & \ d^{(k)}\cdot pT_B(t-\tau)\cdot \sin(2\pi f_m(t-\tau)+\phi)+ \\
&\mathcal{M}(t) + \mathcal{N}(t),\nonumber
\end{split}
\label{Rx_FHSS}
\end{equation}
where $\tau$ is the propagation delay that we are using for calculating the distance, $\mathcal{N}(t)$ is the Gaussian noise, and $\mathcal{M}(t)$ is the multi-path effect, which can be expressed as the following summation:
\begin{eqnarray}\label{Multi-path}
\mathcal{M}(t) = \sum_{i=1}^{N} \alpha_i\cdot s^{(k)}(t-\tau_i),
\end{eqnarray}
where $\alpha_i$ is the attenuation of path $i$ and $\tau_i$ is the time delay of path $i$. As long as we make sure that the hopping speed is faster than the time delay of each path ($\tau_i$), then before the arrival of any of the reflected signals, we already have changed the frequency and different paths do not interfere with the original signal.

By eliminating the multi-path effect using FHSS technology, the received signal would be just the time-delayed transmitted signal plus noise:
\begin{eqnarray}\label{Rx_FHSS_without_multipath}
r^{(k)}  = d^{(k)}\cdot pT_B(t-\tau)\cdot \sin(2\pi f_m(t-\tau)+\phi) + \mathcal{N}(t).
\end{eqnarray}

\begin{figure}
	\centering
	\includegraphics[width=\linewidth]{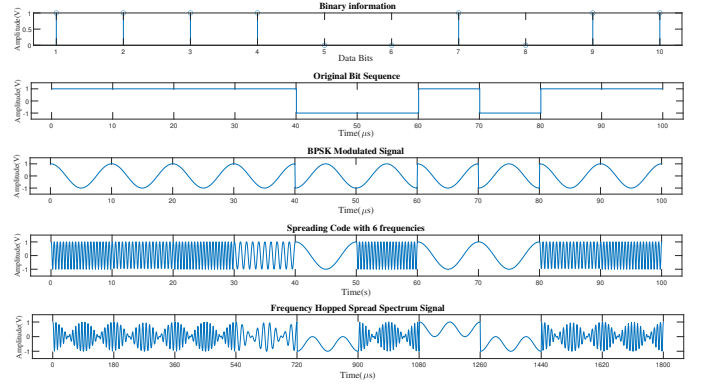} 
	\caption{Illustration on steps of generating the FHSS waveform: (i) Generate a set of random bits; (ii) Convert the bits into rectangular pulses; (iii) Apply the BPSK modulation; (iv) Generate six random sets of spreading codes; (v) Generate the FHSS waveform using the BPSK signals and the spearing codes.}
	\label{Simple_TX}
\end{figure}

\begin{figure}
	\centering
	\includegraphics[width=\linewidth]{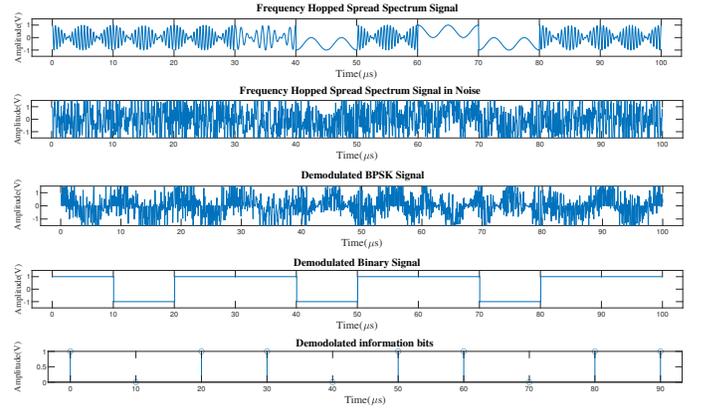} 
	\caption{Illustration of a simple receiving scenario.}
	\label{Simple_RX}
\end{figure}

\begin{figure}
	\centering
	\includegraphics[width=\linewidth]{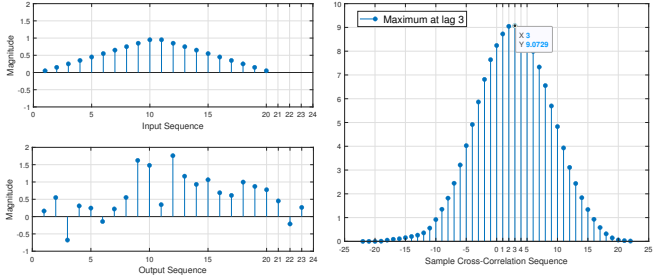} 
	\caption{Simple illustration of how the cross-correlation function works to find the bit at which the peak occurs, which later be translated into the time delay: (i) Input Sequence: Make a random set of data, in this example, we use a triangle set; (ii) Output Sequence: Add Gaussian noise and the delay to the input sequence; (iii) Perform the cross-correlation function and find the delay based on the observed peak.}
	\label{Cross_Correlation}
\end{figure}

Therefore, by performing a  cross-correlation between the received signal and the known transmitted signal (the one without the time delay) and by detecting the sample bit at which the peak occurs, the distance is calculated as follows:
\begin{eqnarray}\label{Distance_Calculation_2}
d = c_{sound} \cdot TOF = c_{sound} \cdot \frac{n_{samples}}{f_s};
\end{eqnarray}
where $n_{samples}$ is the sample number of the maximum peak and $f_s$ is the sampling frequency. 
Fig.~\ref{Simple_TX} depicts a simple scenario of how the transmitter generates the FHSS modulated waveform for transmission. 
Fig.~\ref{Simple_RX} shows how the ultrasound receiver demodulates the FHSS waveform after it has passed through the channel.
Fig.~\ref{Cross_Correlation} shows how the time delay is getting measured at the receiver. Fig.~\ref{Cross_Correlation} is just to depict a clear representation of how the cross-correlation process works; therefore, it uses a simple input and not the actual FHSS waveform that PILOT uses.

For calculating distances using ultrasonic signals, the velocity of ultrasonic signals must be known. As~\cite{Indoor_Acoustic_CDMA} suggested, there are two ways to determine the exact velocity. As a first alternative, the velocity of the sound can be determined from its nominal value at a certain temperature as follows:
\begin{eqnarray}\label{Sound_Speed}
	c_{sound} = 331.3\sqrt{1+\frac{T}{273.15}}.
\end{eqnarray}
Eq.~\ref{Sound_Speed} uses the velocity of the sound at $0 \ ^{\circ}C$ as a reference in order to find the value $c_{sound}$ in ($\frac{m}{s}$) at a temperature $T$ in $^{\circ}C$. The other way to determine the velocity is by measuring the traveling time of the ultrasound signal at a known distance. In our system, we calculated the velocity using both methods. In the first method, a digital thermometer can be deployed to measure the ambient temperature to substitute in the Eq.~\ref{Sound_Speed}. On the other hand, in the second method, the $c_{sound}$ can be constantly updated by locating a reference ultrasound speaker at a known position in the room where it constantly transmits a pilot signal to the ultrasound microphones installed at known locations in the room. 

Besides temperature, humidity and wind are also effective on sound velocity. However, the influence of the humidity is negligible compared to the one of the temperature and intense wind is not applicable in our indoor setup. The effect of indoor airflow due to air-conditioning was investigated in~\cite{Airflow_Affect_1, Airflow_Affect_2}. In order to reduce the influence of wind, Martin et al. in~\cite{Wind_Affect_Reduction} suggested estimating the sound velocity continuously by measuring the travel time for a known distance in proximity and correcting the other measurements accordingly, which is the exact same technique we used as the second method of measuring the velocity of sound.

\begin{figure}
	\centering
	\includegraphics[height=1.2in,width=3.5in,trim={0 14cm 18.5cm 0},clip]{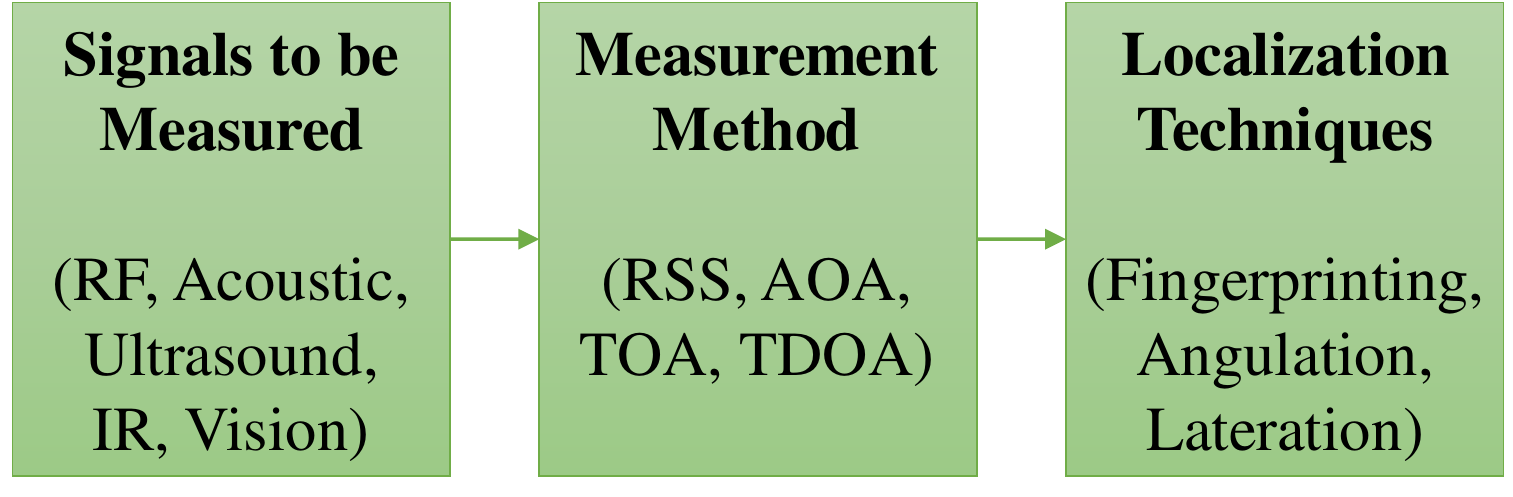} 
	\caption{Ranged-based Location Estimation.}
	\label{Localization_Diagram}
\end{figure}

\section{Enhancing the Robustness of FHSS Ranging} \label{sec:Enhancing the Robustness of FHSS localization and Tracking}
In the previous section, we proposed a method for ranging and estimating the distance between the drone and receiver beacons based on the TOA of ultrasonic signals and for overcoming the multipath, we proposed the FHSS technique. Another approach for keeping track of a drone's movement is by estimating the relative velocity of the drone with respect to each of the receiver beacons. This can be done by measuring the received signals' frequency shift (i.e., Doppler-shift effect) and then estimating the distance. However, the key drawback with using this method alone is that the error can increase over time, and thus, over a prolonged period, this method alone will not provide reliable and accurate tracking. 

However, leveraging a Kalman filter to incorporate the data from both methods (i.e., TOA-based distance estimation and velocity estimation based on frequency shift measurement) can prevent the velocity estimation errors from increasing over time.  In this approach, data collected from each of the aforementioned methods play an essential role in preserving the system's high accuracy location estimation capability over time. On the one hand, velocity estimation based on frequency shift measurements cancels the TOA-based distance estimation measurement error and improves system accuracy. On the other hand, the data obtained from the TOA-based distance estimation at each moment is the primary source of initial data for the final distance calculation, and the method is not merely dependent on the velocity estimate, which makes it unreliable over time. Thus, the combination of the data from both methods keeps the system's distance estimation highly accurate at all times. This approach achieves greater noise robustness and provides more precise localization and monitoring of the drone's movement. This section first describes how PILOT estimates the velocity using change in the frequency and then explains how it integrates distance and velocity estimations in a Kalman filter.

\subsection{Velocity Estimation by Measuring Doppler Shift}
The following equation is being used to express the Doppler shift effect:
\begin{eqnarray}\label{Doppler-shift}
F_s = \frac{\textbf{v}}{c}\cdot F,
\end{eqnarray}
where $F_s$ is the amount of the frequency shift, $\textbf{v}$ is the relative velocity between transmitter and receiver, $c$ is the speed at which the signal propagates ($c$ would denote the speed of light if the signal is an RF signal, and similarly, $c$ would denote the speed of sound if the signal is an acoustic signal), and $F$ is the actual frequency in which the signal is transmitted. Using Eq.~\ref{Doppler-shift}, whether the receiver and transmitter move towards or away from each other can be established by treating $\textbf{v}$ as a vector in the equation.

Several studies have used Doppler shift measurements to estimate the speed and direction of a UAV. For example, in~\cite {Potential_Sound_Based}, to estimate a flying UAV's speed and direction, Shin et al. used Doppler-shift. Nevertheless, depending solely on the Doppler shift has limitations. For instance, it is necessary to know the original location of the target drone to be able to estimate its location continuously at each moment. Moreover, it is almost impossible to achieve high accuracy because of the propagation of the error in the result over time. To overcome these limitations, some methods use Doppler shift measurements as an auxiliary localization tool to improve the efficiency of the primary localization method~\cite{MobiSys_Follow_Me_Drone,Mouse_in_Air,CAT}. PILOT also adopts this strategy.

To estimate the relative velocity between the target drone and each of the receiver beacons fixed throughout the room (where the drone is located), PILOT uses the following procedure: 
\begin{itemize}
\item The ultrasound transmitter system mounted on-board the drone continuously transmits the FHSS waveform at frequency $F = f_m$.
\item After receiving the signal in each of the receiver beacons, PILOT first applies Fast Fourier Transform (FFT) techniques to obtain the frequency content of the received signal and find the peak in the frequency domain and then estimates the frequency shift of the signal, $F_s$, by calculating the difference between the peak frequency of the received signal and $F$.
\item An estimate of the drone's velocity is calculated using Equation~\ref{Doppler-shift}.
\end {itemize}

\subsection{Combining Distance and Velocity Estimation Using a Kalman Filter}
PILOT incorporates the distance estimations obtained by the FHSS-based ranging method and the estimates from Doppler shift-based velocity estimation method. The reason for this approach is to boost the accuracy and robustness of distance estimation performance against noise. There are two ways to combine these two forms of estimates: using a Kalman filter~\cite{MobiSys_Follow_Me_Drone} or through an optimization framework~\cite{CAT}.
In the latter method, we need to devise an optimization framework that consists of both estimates in its objective function. Then, similar to~\cite{CAT}, we can form an optimization framework as follows:
\begin{eqnarray}\label{Optimization_Framework}
	\sum_{i\in[k-n+1...k]}^{max}\sum_{j}^{max}\alpha(|\mathbf{x_i}-\mathbf{c_j}|-|\mathbf{x_0}-\mathbf{c_j}|-d^{i,j}_{FHSS})^2 + \nonumber \\  \sum_{i\in[k-n+2...k]}^{max}\sum_{j}^{max}\beta(|\mathbf{x_i}-\mathbf{c_j}|-|\mathbf{x_{i-1}}-\mathbf{c_j}|-v^{Doppler}_{i-1,j})^2;
\end{eqnarray}
where $k$ is the current processing interval, $n$ is the number of intervals used in the optimization, $\mathbf{x_i}$ denotes the drone's position at the beginning of the i-th interval, $\mathbf{x_0}$ denotes the reference position, $\mathbf{c_j}$ denotes the $j-th$ ultrasound receiver's position in the room, $d^{i,j}_{FHSS}$ denotes the distance change from the reference location with respect to the $j-th$ ultrasound receiver at the $i-th$ interval, $v^{Doppler}_{i,j}$ denotes the relative velocity between the drone and the $j-th$ ultrasound receiver during the $i-th$ interval, $T$ is the interval duration, and $\alpha$ and $\beta$ are the relative weights assigned to the distance and velocity measurements. The only unknown in the optimization is the drone's position over time (i.e., $\mathbf{x_i}$).

The objective reflects the goal of finding a solution $\mathbf{x_i}$, which is the drone's position, that best fits the distances from FHSS ranging and velocities from Doppler shift measurements. The first term reflects that the distance calculated based on the coordinates should match the distance estimated from the FHSS ranging. Similarly, the second term captures that the distance traveled over an interval computed from the coordinates should match the distance derived by multiplying the interval time to the velocity obtained from the Doppler shift measurements. The objective function consists of terms from multiple intervals to improve the accuracy. The formulation in Eq.~\ref{Optimization_Framework} is general and not restricted just to one dimension for tracking the drone on a line, and it can support both two-dimensional and three-dimensional coordinates. In fact, $\mathbf{x_i}$ and $\mathbf{c_j}$ are both vectors whose sizes are determined by the number of dimensions.

Because this optimization problem is non-convex, there is no guarantee of convergence, and it may be slow and computationally expensive. There is some work in the literature to address this issue. Similar to the solution in~\cite{CAT}, some changes to the optimization framework are required to make the problem convex. 
To simplify the objective, we create a new parameter called $D_{i,j}$ that denotes the drone's distances to different ultrasound receiver beacons over time (i.e., replacing $|\mathbf{x_i} \-- \mathbf{c_j}|$ in the objective function with $D_{i,j}$), which is a convex function in terms of $D_{i,j}$. However, not all distances are feasible (i.e., there may not exist coordinates that satisfy the distance constraints). As a result, additional constraints must be derived to enforce feasibility. To begin, a convex relaxation of the original problem must be solved by treating distances as unknowns and replacing feasibility constraints with triangular inequality constraints. Triangular inequality constraints are necessary for the distances to be realizable in a low-dimensional Euclidean space. They are also sufficient in a two-dimensional space to ensure feasibility, but not in a three-dimensional space. As a result, the solution obtained in the first step must be projected into a feasible solution space in the following step. This projection is related to network embedding, which embeds network hosts in a low-dimensional space while preserving their pairwise distances to the greatest extent possible. To rectify this problem, Mao et al.~\cite{CAT} developed an embedding method based on the Alternating Direction Method of Multipliers (ADMM)~\cite{ADMM_ref} to efficiently solve the problem.

One possible benefit of this method compared to the Kalman filter is to combine the Inertial Measurement Unit (IMU) information captured from drone's IMU sensors with the distances from FHSS ranging and velocities from Doppler shift, another term could be added to the objective function, which represents the effect of IMU measurements. However, since PILOT does not depend on the IMU measurements, there is no benefit in using the optimization framework. 

We concluded that, because of its substantially higher computational complexity, the optimization method is insufficient, which would place an enormous burden on the PILOT processing elements. Therefore, PILOT employs the Kalman filter approach. One of our primary objectives is to ensure that the computational complexity and overhead of PILOT's communication are low relative to the state of art. The following paragraphs will provide a brief overview of how PILOT blends distance and velocity estimates using a Kalman filter.  

Let $D_k$ denote the actual distance between the ultrasound transmitter and a receiver beacon in the $k$-th window, $t$ denote the duration, $v_k$ denote the measured Doppler velocity, $n_k$ capture the error in Doppler measurements, $d_k$ denote the measured distance, and $w_k$ denote the distance measurement error. The following equations describe the relationship between these variables: 
\begin{eqnarray}
D_k &=& D_{k-1} + v_k\cdot t + n_k \nonumber \\
d_k &=& D_k + w_k
\end{eqnarray}
As we mentioned before, $d_k$ and $v_k$ are from distance and velocity measurements, respectively. We can utilize the redundancy between these two measurements by using a Kalman filter to decrease the impact of noise and further enhance the accuracy of distance estimation. According to \cite{MobiSys_Follow_Me_Drone}, the optimal distance estimation $\hat{D}_k$ is given by:
\begin{eqnarray}
& \hat{D}_k = \hat{D}_{k-1} +v_k\cdot t + \frac{\hat{p}_{k-1}+q_k}{\hat{p}_{k-1}+q_k+r_k}(d_k-\hat{D}_{k-1}-v_k\cdot t),\nonumber & \\
\end{eqnarray}
where $\hat{p}_k=\frac{r_k(\hat{p}_{k-1}+q_k)}{\hat{p}_{k-1}+q_k+r_k}$, and the variables, $q_k$ and $r_k$, denote the standard deviation for $n_k$ and $w_k$, respectively.
\section{Three-dimensional Localization}\label{sec:Trilateration}
After estimating the distance between an ultrasonic transmitter and a receiver, the next step is to localize the transmitter in three dimensions. Angulation and lateration are the two most well-known localization techniques used to estimate the target object's position based on the angle or distance measurements respectively. As we mentioned in previous sections, due to the challenges with the angle-based measurement methods, PILOT is measuring distance, hence the lateration technique for the final three-dimensional localization of the drone. 
\begin{figure}
	\centering
	\includegraphics[width=\linewidth]{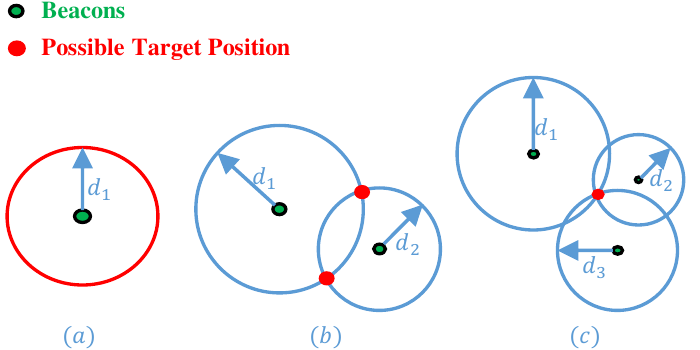} 
	\caption{Graphical representation of the trilateration scheme in two dimensions. It requires at least three beacons for the target drone with an unknown position to localize itself in two dimensions. (a) Target drone has the distance of $d_1$ from the beacon. All the points on the circle with the radius of $d_1$ with the center of the beacon is a possible position for the drone, but no certain answer. (b) The target drone has the distance of $d_1$ to the first beacon and $d_2$ to the second beacon. Two circles intersect in two points, and these two points are the possible position for the target drone; again no certain answer. (c) The target drone has the distance of $d_1$ to the first beacon, $d_2$ to the second beacon, and $d_3$ to the third beacon. The intersection of three circles is just one point which is the position of the target drone.}
	\label{2D_Trilateration}
\end{figure}
To use trilateration for localizing an object in two dimensions, the distances between the object and at least three sources are needed. Similarly, in three-dimensional localization, in order to uniquely localize the target object, we need to measure the distances between the target object and at least four different sources. Fig.~\ref{2D_Trilateration} explains this better. Let's denote the distance between a transmitter and $i$-th receiver as $d_i$. Also, the position of the transmitter is denoted as $[x \ y \ z]^T$ and the position of the $i$-th receiver is denoted as $[x_i \ y_i \ z_i]^T$. Then using trilateration rules, we have:
\begin{eqnarray} \nonumber
	&(x_1-x)^2+(y_1-y)^2+(z_1-z)^2 = d_1^2\\ \nonumber
	&(x_2-x)^2+(y_2-y)^2+(z_2-z)^2 = d_2^2\\ \nonumber
	&\vdots\\
	&(x_n-x)^2+(y_n-y)^2+(z_n-z)^2 = d_n^2 
\end{eqnarray}

After applying some mathematical manipulation, we can then simplify these quadratic equations and write them down in the linear form of $\textbf{A} \textbf{x} = \textbf{b}$ where $\textbf{A}$ and $\textbf{b}$ are equal to:
\vspace{-0.1in}
\begin{small}
	\begin{eqnarray}
		\textbf{A} & = & \begin{bmatrix}
			2(x_n-x_1) & 2(y_n-y_1) & 2(z_n-z_1) \\
			2(x_n-x_2) & 2(y_n-y_2) & 2(z_n-z_2)\\
			\vdots     & \vdots     & \vdots\\
			2(x_n-x_{n-1}) & 2(y_n-y_{n-1}) & 2(z_n-z_{n-1})\\
		\end{bmatrix},\nonumber \\
		\textbf{b} & = & \begin{bmatrix}
			d_1^2 - d_n^2 - x_1^2 -y_1^2 -z_1^2 + x_n^2 + y_n^2 + z_n^2 \\
			d_2^2 - d_n^2 - x_2^2 -y_2^2 -z_2^2 + x_n^2 + y_n^2 + z_2^2 \\
			\vdots \\
			d_{n-1}^2 - d_n^2 - x_{n-1}^2 -y_{n-1}^2 -z_{n-1}^2 + x_n^2 + y_n^2 + z_n^2
		\end{bmatrix}.\nonumber
	\end{eqnarray}
\end{small}

The vector $\textbf{x} = [x \ y \ z]^T$ which includes the coordinate of the object that need to be localized would be: $\textbf{x} = (\textbf{A}^T\textbf{A})^{-1}\textbf{A}^T\textbf{b}$. 

\section{Preliminary Simulation Results}\label{sec:Preliminary Simulations}
This section describes the simulation setup used for preliminary tests followed by the result analysis to evaluate the performance of the enhanced FHSS ranging.
\subsection{Simulation Setup}
\subsubsection{Transmitter}
The transmitter subsystem, which it is the ultrasonic transmitters on-board the drone, generates the desired FHSS signals. We used signals in the frequency range between $25$~KHz and $55$~KHz for two reasons. Firstly, frequencies at or below $20$~KHz need to be avoided since that range would interfere with the frequency range of a human voice. Any overlap would result in degraded performance. In addition, according to the Nyquist theorem, in order to prevent aliasing, the sampling rate must be at least twice the maximum frequency, i.e., if the system operates in the frequency range from $25$~KHz to $55$~KHz, then it requires that the sampling rate should be at least $110$~KHz to avoid aliasing. We do not use frequencies above $55$~KHz because we do not intend to deal with high frequencies to avoid both processing and equipment costs, which means that the sampling rate simply needs to be $110 $KHz or more.
PILOT uses the FHSS waveform. In this waveform, the frequency range is from $25$~KHz to $55$~KHz with $6$ sub-frequency carriers located at $27.5$~KHz, $32.5$~KHz, $37.5$~KHz, $42.5$~KHz, $47.5$~KHz, and $52.5$~KHz and the bandwidth dedicated for each of these sub-carriers is $5$~KHz. A single hop occurs within the transmission time of each data bit, and this hopping rate is fast enough (based on the multi-path analysis of the room using image theory and finding the delay spread of the multi-path signals) to mitigate the effects of multi-path interference. 
We set the sampling frequency ($f_s$) to $340$~KHz; this guarantees not having aliasing and also simplifies our calculations.
For modulation, PILOT uses BPSK (Binary Phased Shift Keying) to take advantage of BPSK's high noise robustness.

In~\cite{ROLATIN}, there is just one ultrasonic speaker on-board the drone, which is transmitting the FHSS waveform for ranging purposes. That setup works based on the assumption that an ultrasonic transmitter has an omni-directional propagation pattern. Although this assumption perfectly works in simulation tests, it will not work in real-life experiments. Unlike most RF transmitters, most ultrasonic transmitters have a relatively narrow-angle radiation beam, i.e., only receivers within the imaginary cone-shape propagation pattern in front of the transmitter can receive the signal. As we mentioned in previous sections, three-dimensional localization requires distances between the transmitter and at least four receivers. Since we put the receivers on the walls of the room, just one of them can receive the transmitted signal from an ultrasonic transmitter on-board the drone, i.e., only the receiver in front of that transmitter can receive the signal. To overcome this issue, we propose the four-leaf clover transmitter setup where the ultrasonic transmitter setup on-board the drone consists of four transmitters facing the front, back, and sides. In this way, at each moment, the same FHSS waveform is transmitted from each of these four transmitters and all the receivers in the room can receive the signal.

\subsubsection{Channel}
We used a Rayleigh channel model in the simulations, which considers the impact of the multi-path fading interference on the transmitted signals. Moreover, we used additive white Gaussian noise (AWGN) to show the impact of the noise in our localization process. Furthermore, we assumed that the movement of the target drone is confined to a rectangular room with dimensions of $5$~m~$\times$~$5$~m~$\times$~$3$~m. This is the typical dimension for indoor office spaces.

\subsubsection{Receiver}
The receiver sub-system, which it is the four receiver beacons located at known positions in the room, cross-correlates the received signal with a reference signal, and then seeks the sample bit that makes the cross-correlation peak. After finding the sample bit in which the cross-correlation peak happens, the receiver sub-system estimates the distance from the drone to each of the receiver beacons using the following relation: 
$d = n_{samples}\times c_{sound}/f_s$, where $n_{samples}$ is the sample number where the maximum cross-correlation occurs and $f_s$ is the sampling frequency. There are four receiver beacons in the room, and the same procedure is used by each of them. 

To achieve the best coverage for any possible drone location in the room, we placed the four receiver beacons in positions to best ensure that at least one receiver beacon is in the line of sight and very close to the corresponding transmitter on-board the drone that is aiming at that receiver. Therefore, the receiver beacons can receive the ultrasound signal with a high signal-to-noise ratio (SNRs) at every drone location in the room. Specifically, the $(x,y,z)$ coordination of the ultrasound receivers in the room is $(2.5,0,1.5)$, $(5,2.5,2.5)$, $(2.5,5,2)$, and $(0,5,3)$ where all the numbers are in meters.

\subsection{Simulation Results}
In this part, we show the results of our preliminary simulations to evaluate the performance of the first two stages of PILOT by measuring the error between the estimated position of the drone and its actual position. We assess the performance of the scheme by benchmarking it with respect to a basic reference scheme that uses \emph{only} FHSS-based distance estimation (before enhancement) to locate a target drone.

For the simulation tests, we ran Monte Carlo with a large enough number of iterations. Moreover, to present the final error, we averaged the error between the actual position of the drone and the estimated position over the entire flight trajectory. We ran the simulation over many different trajectories that were generated randomly. This ensures that the performance evaluation is valid over any possible drone's trajectory, and not just for some specific scenarios. To showcase the result, we randomly chose seven of the flight trajectories and demonstrated the performance of PILOT on them. The placement of the ultrasound receivers remained the same for all of the trajectories.

\begin{figure}
	\centering
	\includegraphics[width=\linewidth]{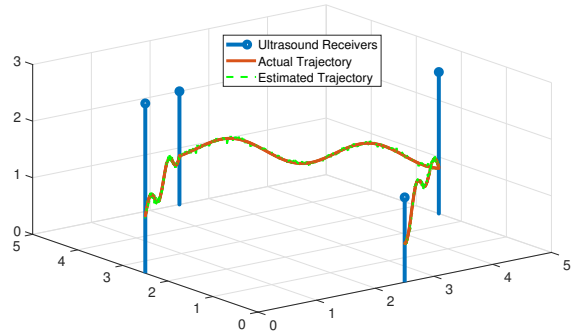} 
	\caption{Representation of the drone's estimated trajectory in comparison with the drone's actual trajectory. The placement of the four ultrasound receivers is also depicted in this figure.}
	\label{Trajectory}
\end{figure}

In Fig.~\ref{Trajectory}, we show a drone's actual trajectory as well as the estimated trajectory using FHSS distance estimation and Doppler shift velocity estimation incorporated in the Kalman filter (the first two stages of the PILOT). This is one of the drone's random flight trajectories in the simulation tests. The locations of the ultrasonic receivers are also indicated in this figure. The actual and estimated trajectories seem to overlap perfectly because, relative to the room's dimensions in which the drone's movement is confined, the localization estimation error is significantly small.

\begin{figure}
	\centering
	\includegraphics[width=\linewidth]{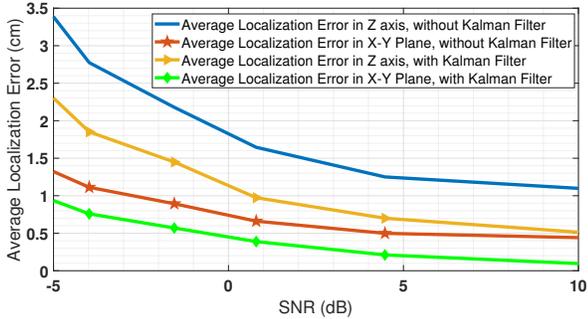} 
	\caption{Average localization error (cm) vs. SNR (dB) for $X-Y$ plane and $Z$-axis for PILOT's first stage only (as a benchmark) and PILOT's complete first two stages together (FHSS ranging and Doppler shift velocity estimation incorporated in the Kalman filter).}
	\label{SNR}
\end{figure}

Fig.~\ref{SNR} illustrates the relationship between the localization performance and the signal-to-noise ratio (SNR) of the received signal at each of the ultrasound receivers. As predicted, the localization error is inversely proportional to the signal's SNR value. Moreover, as is shown in this figure, deploying the second stage of the PILOT improves the localization accuracy both in the $X-Y$ plane and in the $Z$-axis. Finally, in the figure, we note that the $Z$-axis localization error is much greater than that of the $X-Y$ plane. We will thoroughly investigate this issue in the next section and show that the reason behind this is the relative geometry between the transmitter setup and the receivers, i.e., the vertical factor of the relative geometry is much higher than the horizontal factor.

\begin{figure}
	\centering
	\includegraphics[width=\linewidth]{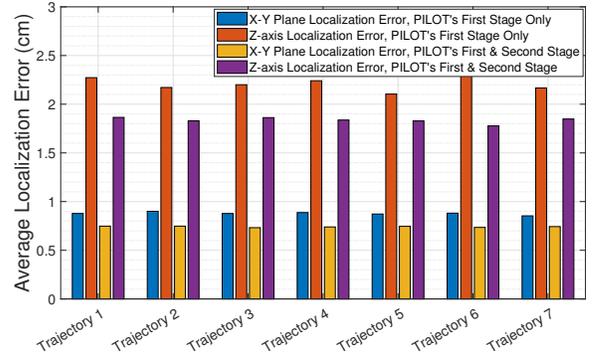} 
	\caption{Average localization error comparison between PILOT's first stage (FHSS ranging only) and PILOT's first two stages (FHSS ranging and Doppler shift velocity estimation incorporated in the Kalman filter) for both the $Z$-axis and $X-Y$ plane.}
	\label{different_Trajectories}
\end{figure}

Fig.~\ref{different_Trajectories} compares the localization error between the first stage of the PILOT (FHSS ranging only) which as we mentioned before, is the benchmark, and the complete first two stages of PILOT (FHSS ranging and Doppler shift velocity estimation incorporated in the Kalman filter). As is shown in the figure, the $X-Y$ plane localization error is plotted separately from the $Z$-axis to show the massive difference between them and justify the need to implement the third stage of PILOT. As can be seen in this figure, in all the seven random drone flight trajectories, the second stage of PILOT slightly improves the localization error both for the $X-Y$ plane and the $Z$-axis; however, the $Z$-axis localization error is still more than twice of which in the $X-Y$ plane.

\section{Height Estimation Improvement} \label{sec:Height Enhancement}
The first two stages of PILOT provide high accuracy localization in the $X-Y$ plane; however, the $Z$-axis localization error still needs to be improved because it is much greater than that of the $X$ or $Y$ axis. Generally, the localization error for the ranging-based localization methods originates from two sources, first the error in estimating the distance between the target and each beacon, known as the ranging error, and second the error arising from the relative geometry between the target and all of the beacons. The reason for having a more significant error in the $Z$-axis compared to the $X-Y$ plane is the relative geometry between the transmitter setup on-board the target drone and the receiver beacons in the room.

Thus far, we have shown how the first two stages of PILOT abate the ranging error by deploying the FHSS communication scheme for the distance estimation task and making the scheme robust against noise and the indoor multi-path fading effect. In this section, we first thoroughly investigate the reason behind having a bad $Z$-axis estimation and then show how PILOT provides a solution to lessen the $Z$-axis estimation error, leading to a better overall three-dimensional accuracy. To the best of our knowledge, PILOT is the first scheme that provides solutions to mitigate both the ranging and the geometry-related errors and proposes highly accurate localization for drones in indoor environments.

The Cramer-Rao Bound (CRB), which is the lower bound on the position variance that can be obtained using an unbiased location estimator~\cite{CMU}, is a useful metric for evaluating the localization accuracy. In~\cite{CMU}, Rajagopal showed that for a two-dimensional trilateration system with an unbiased estimator, under the assumption that the range measurements are independent and have zero-mean additive Gaussian noise with constant variance $\sigma^2_r$, the CRB variance of the positional error $\sigma^2(r)$ at position $r$, as defined by $\sigma^2(r) = \sigma^2_x(r) + \sigma^2_y(r)$ is given by:
\begin{eqnarray}
	\sigma(r) = \sigma_r \times \sqrt{\frac{N_b}{\sum_{k=1}^{N_b-1}\sum_{j=k+1}^{N_b}C_{kj}}}, \nonumber
\end{eqnarray}
where $N_b$ is the number of beacons, $C_{kj} = |\sin(\theta_k - \theta_j)|$, $\theta_k$ is the angle between $b_k$ and $r$, and $b_k$ is the $k$-th beacon.

This illustrates that the error of localization is a multiplication of the error of range measurement with another variable that is a function of the number of beacons and the angle between the beacons and the target object. This function is called Geometric Dilution of Precision (GDOP) and we have: $\sigma(r) = \sigma_r \times GDOP$. Since CRB is directly proportional to the GDOP, GDOP can be used as a reasonable guide for measuring the accuracy of the localization~\cite{CMU,Relative_location_estimation,Cellular_Mobile_Estimation,Cooperative_localization}.

In general, for three-dimensional localization of an object at $[x \ y \ z]^T$ using ultrasound beacons, we have:
\begin{eqnarray}
	\sqrt{Var(x)+Var(y)+Var(z)+Var(c\tau)} = GDOP \cdot \sigma_r, \nonumber
\end{eqnarray}
where $c$ here is the speed of sound and $\tau$ is the receiver's clock offset. Since we have synchronization between the transmitter and the ultrasound receiver beacons in our work, then the timing offset is considered to be zero; therefore:
\begin{eqnarray} \label{GDOP_1}
	GDOP = \sqrt{\frac{\sigma^2_x+\sigma^2_y+\sigma^2_z}{\sigma^2_r}}.
\end{eqnarray}

The distance between the drone and each of the beacons is calculated from the following:
\begin{eqnarray} \label{r}
	r_i = \sqrt{(x-x_i)^2 + (y-y_i)^2 + (z-z_i)^2},
\end{eqnarray}
where, as discussed earlier, $[x \ y \ z]^T$ denote the drone's position and $[x_i \ y_i \ z_i]^T$ denote the position of the $i$-th receiver beacon.
The exact $r_i$ is not known due to the ranging measurement error and that causes errors in the Eq.~\ref{r} solution for $[x \ y \ z]^T$. Similar to~\cite{xDOP_Formulas}, we take the differential of Eq.~\ref{r} and disregard terms beyond first order to find a relationship between the solution errors and the ranging errors between the drone and each of the ultrasound receiver beacons in the room:
\begin{eqnarray}
	\Delta r_i = \frac{\Delta x(x-x_i) + \Delta y(y-y_i) + \Delta z(z-z_i)}{\sqrt{(x-x_i)^2 + (y-y_i)^2 + (z-z_i)^2}} \nonumber \\
	= \Delta x \cos \alpha_i + \Delta y \cos \beta_i + \Delta z \cos \gamma_i, \nonumber
\end{eqnarray} 

\begin{figure}
	\centering
	\includegraphics[width=\linewidth]{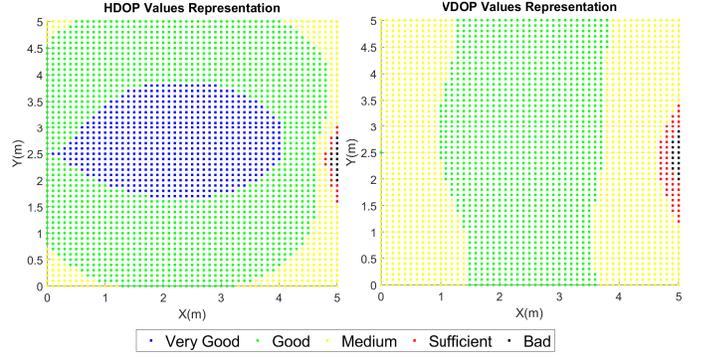} 
	\caption{Graphical representation of the HDOP and VDOP values in the room.}
	\label{fig:xDOP}
\end{figure}

Let $\mathbf{\Delta X} = [\Delta x \ \Delta y \ \Delta z]^T$ be the position error vector and $\mathbf{\Delta R} = [\Delta r_1 \cdots \Delta r_n]^T$ be the target range error vector. Then we can define matrix $\textbf{C}$ as:
\begin{eqnarray}
	\textbf{C} & = & \begin{bmatrix}
		c^1_1 & c^1_2 & c^1_3 \\
		\vdots     & \vdots     & \vdots\\
		c^n_1 & c^n_2 & c^n_3 \\
	\end{bmatrix},\nonumber
\end{eqnarray}
where $[c^i_1 \ c^i_2 \ c^i_3] = [\cos \alpha_i \cos \beta_i \cos \gamma_i]$. Now we can write $\mathbf{\Delta R} = \textbf{C} \mathbf{\Delta X}$ and then we have $\mathbf{\Delta X} = (\textbf{C}^T\textbf{C})^{-1} \textbf{C}^T \mathbf{\Delta R}$. We know that:
\begin{eqnarray} \label{GDOP_2}
	\textbf{Cov}(\mathbf{\Delta X}) = \textbf{E}(\mathbf{\Delta X}\mathbf{\Delta X}^T) =   
	\begin{bmatrix}
		\sigma^2_x & \sigma_{xy} & \sigma_{xz} \\
		\sigma_{yx} & \sigma^2_y & \sigma_{yz}  \\
		\sigma_{zx} & \sigma_{zy} & \sigma^2_z  \\
	\end{bmatrix}. 
\end{eqnarray}
If we assume that Var($r_i$) = $\sigma^2_r$ and that the errors $\Delta r_i$ are uncorrelated, then:
\begin{eqnarray} \label{GDOP_3}
	\textbf{E}(\mathbf{\Delta X}\mathbf{\Delta X}^T) = \textbf{E}(((\textbf{C}^T\textbf{C})^{-1} \textbf{C}^T \mathbf{\Delta R})((\textbf{C}^T\textbf{C})^{-1} \textbf{C}^T \mathbf{\Delta R})^T) \nonumber \\
	= (\textbf{C}^T\textbf{C})^{-1} \textbf{C}^T \textbf{E}(\mathbf{\Delta R}\mathbf{\Delta R}^T) ((\textbf{C}^T\textbf{C})^{-1} \textbf{C}^T)^T \nonumber \\
	= (\textbf{C}^T\textbf{C})^{-1} \textbf{C}^T \textbf{C} (\textbf{C}\textbf{C}^T)^{-1} \sigma^2_r = (\textbf{C}^T\textbf{C})^{-1} \sigma^2_r. \nonumber
\end{eqnarray}
Eq.~\ref{GDOP_1}, Eq.~\ref{GDOP_2} and the above result show that the $(\textbf{C}^T\textbf{C})^{-1}$ diagonal elements can be used to determine the GDOP. GDOP consists of Vertical Dilution of Precision (VDOP) and Horizontal Dilution of Precision (HDOP), i.e., $GDOP = \sqrt{{HDOP}^2+{VDOP}^2}$ where $HDOP = \sqrt{\sigma_x^2+\sigma_y^2}$ is the effect of the relative geometry between transmitter setup and receivers on the $X-Y$ plane's estimation accuracy and $VDOP = \sqrt{\sigma_z^2}$, on the other hand, indicates the impact of geometry on the $Z$-axis estimation. This explains how the $X-Y$ plane localization error can be different from the one for the $Z$-axis. The evaluation of GDOP values is shown in Table~\ref{table:GDOP} \cite{Analysis_Accuracy_of_IPS},\cite{Bad_Geometry_Influence}. In our problem, after measuring the GDOP, HDOP, and VDOP for all the points in the room, we saw that the average of HDOP values over all the points is placed in the "very good" category of the Table~\ref{table:GDOP} whereas the average for VDOP is in the "good" category. This is the reason for having a worse average $Z$-axis estimation error in comparison with the $X-Y$ average estimation error. Fig.~\ref{fig:xDOP} shows a color representation of the HDOP and GDOP values in the room. In the following, we propose our solution to solve this problem.

\begin{table}
	\caption{Evaluation of GDOP Values}
	\label{table:GDOP}
	\begin{center}
		\begin{tabular}{ |c | c | }
			\hline
			\textbf{GDOP Values} & \textbf{Evaluation of the geometry of the beacons} \\
			\hline
			$<1$ & Measurements error or redundancy \\
			\hline
			$1$ & Ideal \\
			\hline
			$1-2$ & Very Good  \\
			\hline
			$2-5$ & Good \\
			\hline
			$5-10$ & Medium \\
			\hline
			$10-20$ & Sufficient  \\
			\hline
			$>20$ & Bad \\
			\hline
		\end{tabular}
	\end{center}
\end{table}

We use a separate ultrasonic transceiver installed on-board the drone to estimate the height continuously, then using a filter, we incorporate this measurement with the $Z$-axis estimation that is already available from the previous steps of PILOT. This significantly improves the overall $Z$-axis estimation accuracy and compensates for having a higher average VDOP compared to the average HDOP.

We place the ultrasonic transceiver facing upward on-board the drone and by measuring the time of flight of the ultrasonic signal emitted from this sensor, after it is reflected from the ceiling, we can determine the distance between the drone and the ceiling. Then, we find the drone's height at each moment by subtracting this result from the room's height. The channel between the drone and the ceiling is typically more reliable than the one between the drone and the floor, since there are usually no error-inducing artifacts between the drone and the ceiling. Following shows the height estimation using this extra ultrasonic transceiver:
\begin{eqnarray}
	h' = \ c_{sound}\cdot t/2 \ ; \ h_{drone} = \ H - h', \nonumber
\end{eqnarray}
where $h'$ is the distance between the drone and the ceiling, $t$ is the total time that takes the signal to travel from ultrasonic transceiver on-board the drone, hit the ceiling, reflecting, and is received in the ultrasonic transceiver on-board the drone, $H$ is the room's height, and $h_{drone}$ is the estimation for drone's height. The revised final location estimation of the drone is then: $\textbf{x} = [x, \ y, \ (w_1\cdot z + w_2\cdot h_{drone})]^T$, where $w_1$ and $w_2$ are the weights assigned in the filter to incorporate the estimated height from the first two stages of PILOT and the third stage respectively in order to calculate the final revised height for the drone.

The complete procedure that PILOT employs to localize a drone in a GPS-denied environment is illustrated below:

\noindent \textbf{$\bullet$ Drone Station:}
\begin{enumerate}
	\item Four ultrasonic transmitters continuously transmit FHSS signals for distance and velocity estimation.
	\item One ultrasonic transceiver continuously measures the drone's height separately.
\end{enumerate}
\textbf{$\bullet$ Receivers:} 
\begin{enumerate}
	\item \textbf{Time Domain Analysis:} Estimate the TOA of the received FHSS signal using the cross-correlation technique and compute the distance: 
	\begin{eqnarray}
		d = \frac{n_{samples}}{f_s}\cdot c_{sound}; \nonumber
	\end{eqnarray}
	\item \textbf{Frequency Domain Analysis:} Estimate the shift in the frequency of the received FHSS signal using the FFT technique and compute the velocity:
	\begin{eqnarray}
		\textbf{v} = c\cdot \frac{F_s}{F}; \nonumber
	\end{eqnarray}
	\item \textbf{Kalman Filter:} Combine the calculated distance and velocity to smooth out the result and find the final distance between the drone and each of the receivers:
	\begin{eqnarray}
		D_k &=& D_{k-1} + v_k\cdot t + n_k, \nonumber \\
		d_k &=& D_k + w_k, \nonumber
	\end{eqnarray}
	\small
	\begin{eqnarray}
		&\hat{D}_k = \hat{D}_{k-1} +v_k\cdot t + \frac{\hat{p}_{k-1}+q_k}{\hat{p}_{k-1}+q_k+r_k}(d_k-\hat{D}_{k-1}-v_k\cdot t);\nonumber
	\end{eqnarray}
\end{enumerate}
\textbf{$\bullet$ Center Station:}
\begin{enumerate}
	\item 3D Localization of drone by applying Trilateration technique on estimated distances between the drone and receivers: 
	\begin{eqnarray}
		\textbf{x} = [x \ y \ z]^T = (\textbf{A}^T\textbf{A})^{-1}\textbf{A}^T\textbf{b}; \nonumber
	\end{eqnarray}
	\item Combine the additional height estimation received from the ultrasonic transceiver on-board the drone with the calculated 3D location to improve the Z-axis estimation:
	\begin{eqnarray}
		\textbf{x} = [x, \ y, \ (w_1\cdot z + w_2\cdot h_{drone})]^T; \nonumber
	\end{eqnarray}
	\item Send the navigation command to the drone controller.
\end{enumerate}

\begin{figure}
	\centering
	\includegraphics[width=0.7\linewidth]{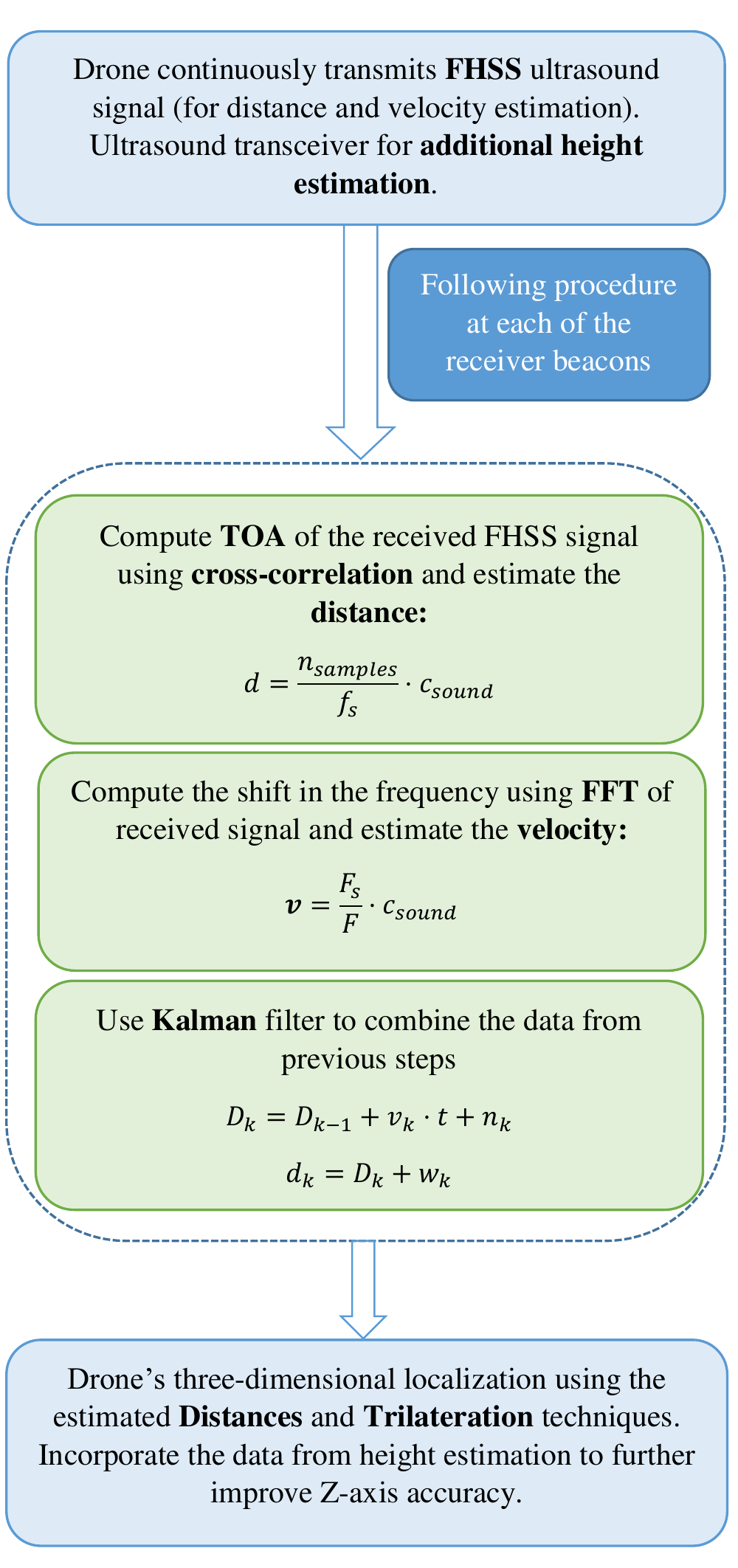} 
	\caption{PILOT's procedure for three-dimensional localization.}
	\label{FlowChart}
\end{figure}

As a graphical summary of the PILOT procedure, Fig.~\ref{FlowChart} represents the scheme that PILOT employs to localize the drone in three-dimensional space accurately.

\section{Experimental Testbed $\&$ Evaluation} \label{sec:Experiments}
The performance of PILOT was comprehensively assessed using real-life experimental tests coupled with simulations in MATLAB. In this section, first, we provide details of the experimental testbed in~\ref{subsec:Experimental Testbed}, and then showcase the evaluation results in~\ref{subsec:Evaluation Results}. 

\subsection{Experimental Testbed}\label{subsec:Experimental Testbed}
\begin{figure}
	\centering
	\includegraphics[height=1.4in,width=2.65in,trim={3cm 21cm 19cm 1cm},clip]{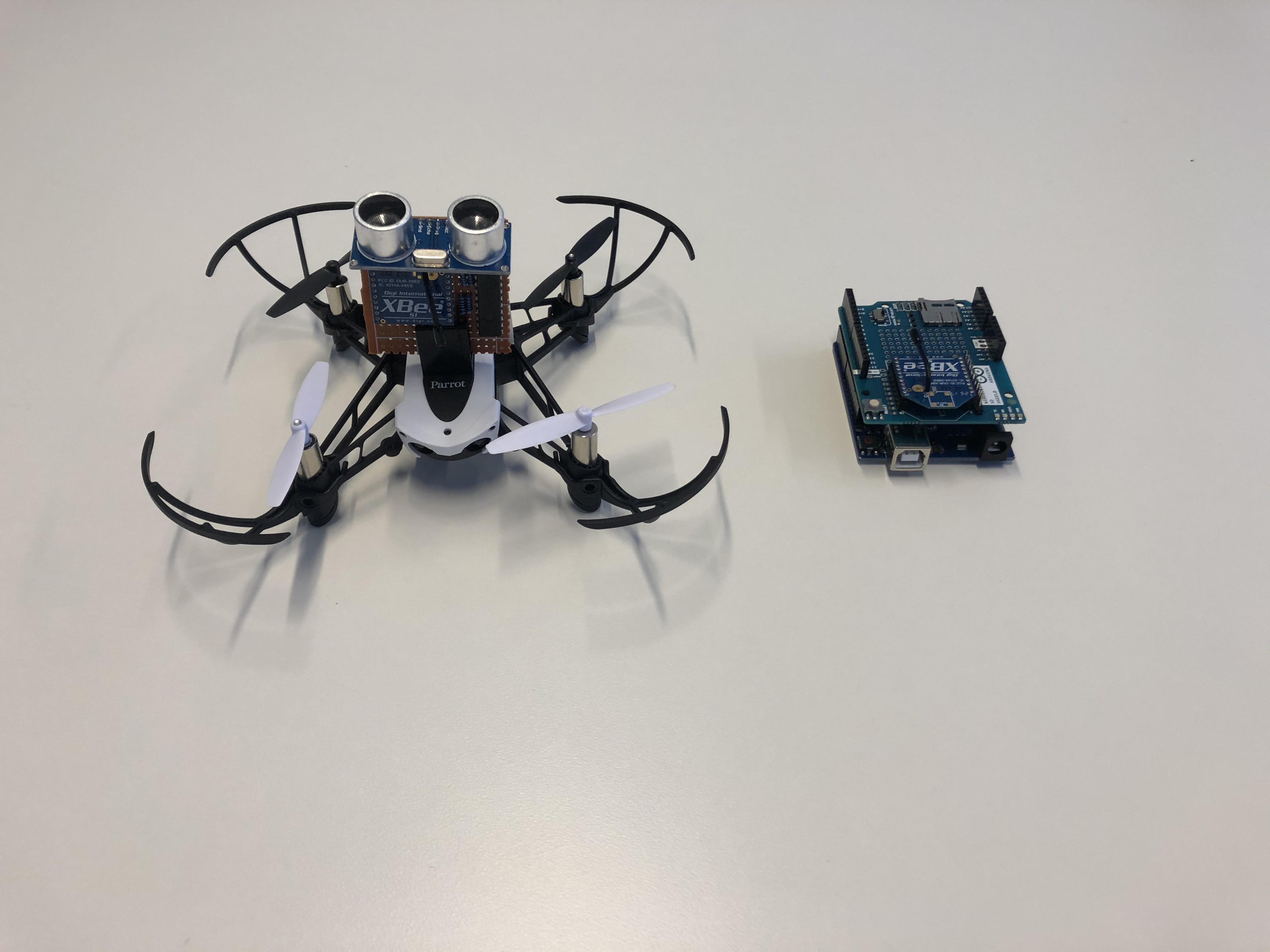} 
	\caption{Parrot Mambo Drone equipped with the ultrasound transceiver system, which has the capability of wireless signal transmission on the left side, and the signal reception on the right side.}
	\label{testbed}
\end{figure}

The experimental test setup consists of two stations, as shown in Fig.~\ref{testbed}. First, which is the left part in the figure, is the drone and the system on-board it. Second, which is the right side of the figure, is the ground control station that helps input the transmitted data into the MATLAB program running on a Dell XPS~$15$ laptop. A Parrot Mambo Drone is the drone used for the experiment. It is a low-cost, off-the-shelf, and ultra-light drone suitable for indoor experiments and has the capacity to hold some light loads as well. The designed system mounted on-board the drone consists of an Arduino Uno micro-controller connected to an $HC$-$SR04$ sensor and a XBee S1 module. The $HC$-$SR04$ sensor is for ultrasonic distance measurement purposes and the XBee S1 module is for wireless communication with the ground controller. In the ground control unit, another Arduino Uno micro-controller connected to a XBee S1 receives the data and transfers it into the MATLAB program running on the laptop. All the experiments were conducted in a hallway inside the building with dimensions $5$~m$~\times$~$5$~m~$\times$~$3$~m.

\subsection{Results $\&$ Evaluation}\label{subsec:Evaluation Results}
In this part, we show the results of our simulations and experiments to evaluate the performance of PILOT by measuring the error between the estimated position of the drone and its actual position. We assess the performance of the scheme by benchmarking it with respect to a basic reference scheme that uses \emph{only} FHSS-based distance estimation (PILO's first stage) to locate a target drone.

For the simulation tests, we used the same setup and configuration as the preliminary tests. Moreover, to present the final error, we averaged the error between the actual position of the drone and the estimated position over the entire flight trajectory. 
For the experimental tests, we flew the Parrot Mambo drone equipped with our system over different trajectories in the room. Having multiple random different flight trajectories ensures that the results are valid for any trajectories and not just restricted to some specific ones. To showcase the evaluation results, we chose five random flight trajectories and presented the result for them. In each of these five random trajectories, we made the route in the room before flying the drone and used them as the ground truth, i.e., we compared the estimated location at each moment with these actual ground truth paths.

\begin{figure}
	\centering
	\includegraphics[width=\linewidth]{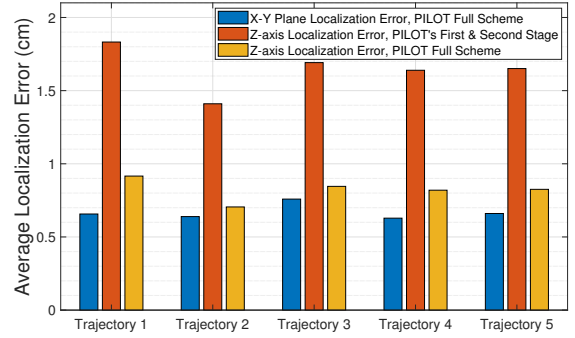} 
	\caption{Evaluating the accuracy of $Z$-axis estimation before and after deploying the third stage and comparison between PILOT's overall $X-Y$ plane estimation accuracy with PILOT's $Z$-axis estimation accuracy.}
	\label{Experiment_1}
\end{figure}

As is seen in Fig.~\ref{Experiment_1}, a comparison between the $X-Y$ plane localization error and the $Z$-axis localization error before deploying the third stage of PILOT shows the significant difference between them and justifies the necessity of deployment of the third stage of PILOT. Moreover, as is shown in this figure, the third stage of the PILOT performs as expected and it improves the $Z$-axis localization error significantly by constantly transmitting the measured data from the supplementary ultrasound transceiver on-board the drone ($HC$-$SR04$) using the XBee S1 wireless module to the receiver module connected to the Dell XPS $15$ laptop and combines this information with current $Z$-axis estimation in a filter.

\begin{figure}
	\centering
	\includegraphics[width=\linewidth]{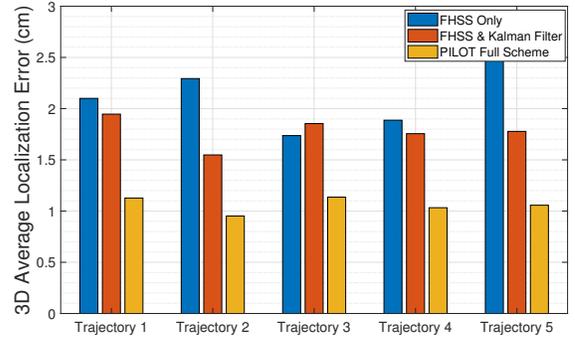} 
	\caption{Evaluating the performance of PILOT: overall three-dimensional localization accuracy.}
	\label{Experiment_Full}
\end{figure}

In Fig.~\ref{Experiment_Full}, we compare the performance of PILOT with that of the benchmark scheme (which relies only on FHSS-based distance estimation to localize a target drone) in terms of the overall three-dimensional localization error. The average value of three-dimensional localization error for PILOT is less than $1.2$~cm. As shown in the figure, the benchmark scheme’s localization error is more than twice that of PILOT. The benchmark scheme focuses on mitigating ranging-based error by deploying the FHSS communication scheme for localization. Other localization schemes proposed in the literature try to improve the localization accuracy by proposing their own techniques to mitigate the ranging-based error. However, PILOT considers both the ranging error and the error due to the relative geometry between the transmitter and receivers and proposes solutions to mitigate both of them and further improve localization accuracy.

PILOT achieves significant improvement in comparison with other drone localization schemes in the literature~\cite{MobiSys_Follow_Me_Drone,Spread_Spectrum_and_MEMS,Robust_Broadband,ROLATIN,Quadrotor_Ultrasonic_Localization,Spread_Spectrum_Ultrasound_and_Time-of-Flight_Cameras}. For instance, in~\cite{ROLATIN}, Famili et al. failed to solve the $Z$-axis estimation error and their scheme had a much greater $Z$-axis estimation error compared to the $X-Y$ plane estimation error. The scheme proposed by Segers et al.~\cite{Spread_Spectrum_and_MEMS} incurs an error of $2$~cm or greater in terms of localization error just for the $X-Y$ plane. Their scheme is merely for two-dimensional localization and they do not consider the complete three-dimensional localization. In~\cite{MobiSys_Follow_Me_Drone}, Mao et al. proposed an FMCW method to overcome the impact of interference and multi-path; however, their system was designed to track the drone on one line, just a single dimension, whereas PILOT proposes a three-dimensional localization. As another example, in~\cite{Quadrotor_Ultrasonic_Localization}, O'Keefe et al. proposed a scheme for three-dimensional localization of drones which incurs an average error of $5.2$~cm which is approximately five times worse than what PILOT provides.

\section{Conclusions} \label{sec:Conclusion}
In this paper, we proposed PILOT, a three-dimensional localization scheme for drones in GPS-denied environments. PILOT takes advantage of the beneficial features of ultrasonic signals and develops a three-stage system to accurately estimate the drone's position in three dimensions. In the first stage, PILOT uses an FHSS-based ranging technique to estimate the distance between the drone and each receiver beacon. The FHSS waveform guarantees robustness against noise and indoor multi-path fading. In the second stage, PILOT first estimates the relative velocity between the drone and each receiver beacon using the Doppler-shift effect. Then, it designs a Kalman filter for estimating the final distance by combining the Doppler shift-based velocity estimation with FHSS-based distance estimation to mitigate the error. In the third stage, by providing a separate height estimation using an additional ultrasonic sensor, PILOT improves the Z-axis estimation error due to the relative geometry between the transmitters and receivers. Conducting thorough simulation tests coupled with real-life experiments to evaluate PILOT's performance shows that PILOT achieves higher localization accuracy compared to the schemes proposed in the literature.     


\footnotesize
\bibliographystyle{IEEEtran}
\bibliography{ref_PILOT}

\end{document}